\documentclass[10pt,twocolumn,letterpaper]{article}

\usepackage{cvpr}
\usepackage{times}
\usepackage{epsfig}
\usepackage{graphicx}
\usepackage{amsmath}
\usepackage{amssymb}

\usepackage{bm}
\usepackage{algorithm}
\usepackage[noend]{algpseudocode}
\usepackage{booktabs}
\usepackage{xcolor}

\usepackage[breaklinks=true,bookmarks=false]{hyperref}

\newcommand{\train}{\mathcal{D}}
\newcommand{\valid}{\mathcal{D_{\mathrm{v}}}}
\newcommand{\test}{\mathcal{D_{\mathrm{test}}}}

\def\sE{{\mathbb{E}}}

\def\vone{{\bm{1}}}
\def\vtheta{{\bm{\theta}}}
\def\vg{{\bm{g}}}
\def\vx{{\bm{x}}}
\def\vy{{\bm{y}}}
\def\vz{{\bm{z}}}

\def\mC{{\bm{C}}}
\def\mG{{\bm{G}}}
\def\mI{{\bm{I}}}

\def\mR{{\bm{R}}}

\def\mZ{{\bm{Z}}}

\def\emR{{R}}

\DeclareMathAlphabet{\mathsfit}{\encodingdefault}{\sfdefault}{m}{sl}
\SetMathAlphabet{\mathsfit}{bold}{\encodingdefault}{\sfdefault}{bx}{n}
\newcommand{\tens}[1]{\bm{\mathsfit{#1}}}

\def\tZ{{\tens{Z}}}

\def\sM{{\mathbb{M}}}
\def\sN{{\mathbb{N}}}

\DeclareMathOperator*{\argmax}{arg\,max}
\DeclareMathOperator*{\argmin}{arg\,min}
\DeclareMathOperator{\tr}{tr}

\cvprfinalcopy 

\def\cvprPaperID{2499} 

\begin{document}

\title{Deep Active Learning for Biased Datasets via Fisher Kernel Self-Supervision}

\author{
	Denis Gudovskiy\\
	Panasonic $\beta$ AI Lab\\
	{\tt\small denis.gudovskiy@us.panasonic.com}
	\and
	Alec Hodgkinson\\
	Panasonic $\beta$ AI Lab\\
	{\tt\small alec.hodgkinson@us.panasonic.com}
	\and
	Takuya Yamaguchi\\
	Panasonic AI Solutions Center\\
	{\tt\small yamaguchi.takuya2015@jp.panasonic.com}
	\and
	Sotaro Tsukizawa\\
	Panasonic AI Solutions Center\\
	{\tt\small tsukizawa.sotaro@jp.panasonic.com}
}


\maketitle
\begin{abstract}	
Active learning (AL) aims to minimize labeling efforts for data-demanding deep neural networks (DNNs) by selecting the most representative data points for annotation. However, currently used methods are ill-equipped to deal with biased data. The main motivation of this paper is to consider a realistic setting for pool-based semi-supervised AL, where the unlabeled collection of train data is biased. We theoretically derive an optimal acquisition function for AL in this setting. It can be formulated as distribution shift minimization between unlabeled train data and weakly-labeled validation dataset. To implement such acquisition function, we propose a low-complexity method for feature density matching using self-supervised Fisher kernel (FK) as well as several novel pseudo-label estimators. Our FK-based method outperforms state-of-the-art methods on MNIST, SVHN, and ImageNet classification while requiring only $1/10$th of processing. The conducted experiments show at least 40\% drop in labeling efforts for the biased class-imbalanced data compared to existing methods\footnote{\href{https://github.com/gudovskiy/al-fk-self-supervision}{Our code is available at github.com/gudovskiy/al-fk-self-supervision}}.
\end{abstract}

\section{Introduction}
\label{sec:intro}
\begin{figure*}[t]
	\centering
	\includegraphics[width=0.85\textwidth]{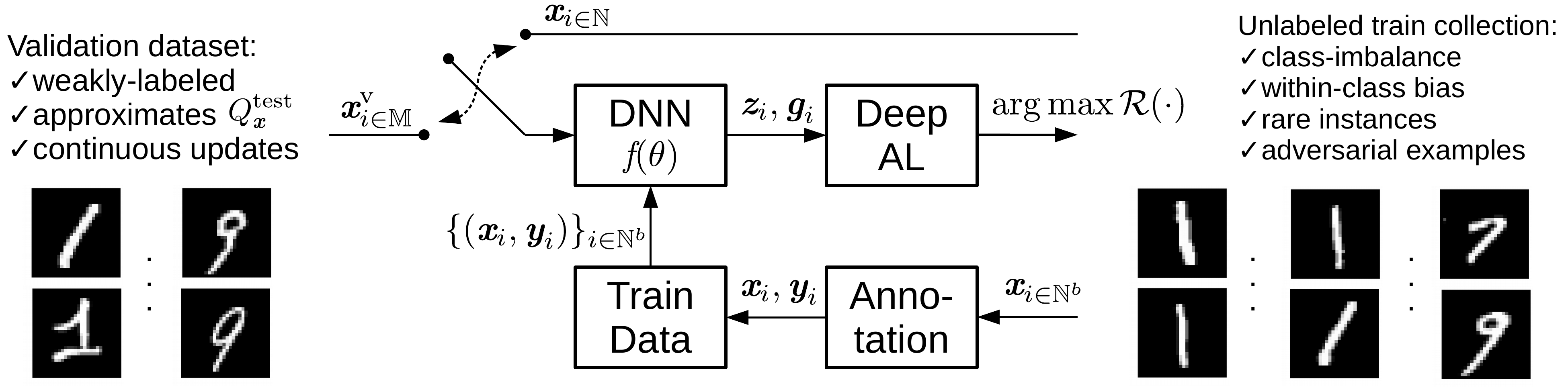}
	\caption{General setup for our semi-supervised AL: validation dataset is selected to approximate test data and can be continuously updated by the newly discovered misclassified examples. Unlabeled collection of train data is subject to the specified distortions. AL algorithm finds relevant train examples for annotation by maximizing acquisition function $\mathcal{R}(\cdot)$ every $b$th iteration.}
	\label{fig:0}
\end{figure*}

Active learning (AL) algorithms aim to minimize the number of expensive labels for supervised training of deep neural networks (DNNs) by selecting a subset of relevant examples from a large unlabeled collection of data~\cite{lewis} as sketched in Figure~\ref{fig:0}. The subset is annotated by an oracle in semi-supervised setting and added to the training dataset in a single \textit{pool} or, more often, in an iterative fashion. The goal is to maximize prediction accuracy while minimizing the pool size and number of iterations.

\begin{figure}[t]
	\centering
	\includegraphics[width=0.98\columnwidth]{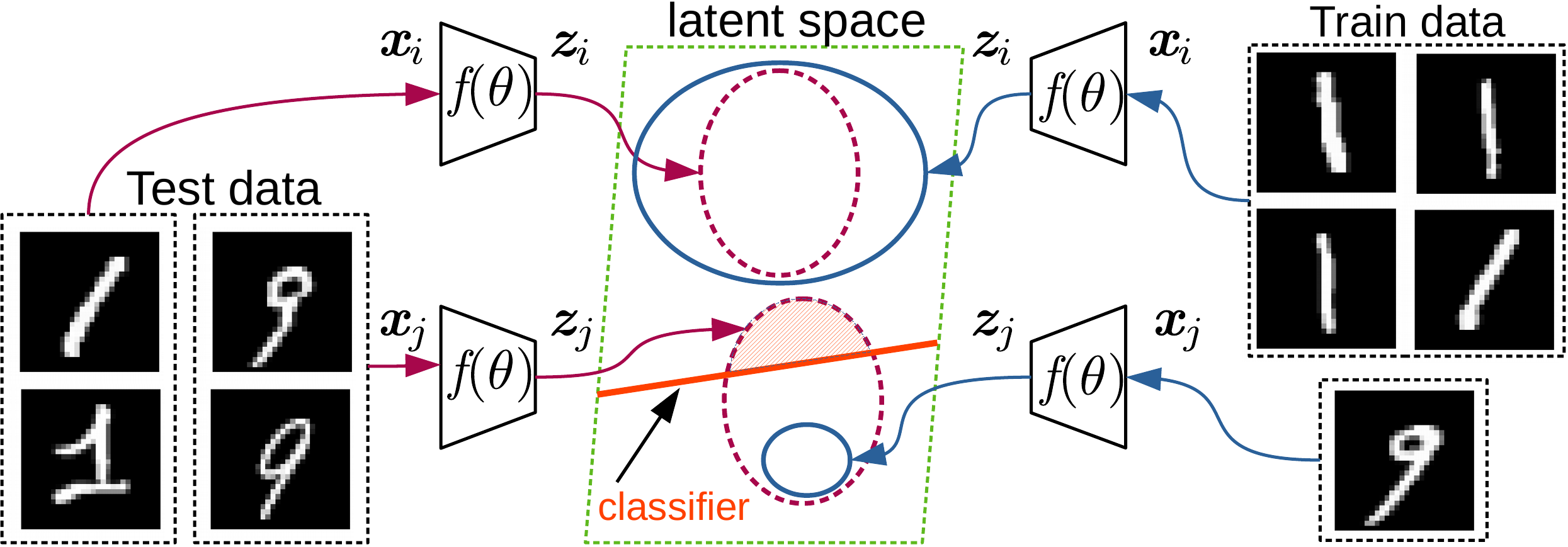}
	\caption{Problem statement for AL with biased data: distribution of unlabeled train data is not aligned with the test data. As a result, prior methods select examples from another distribution and the learned classifier $f(\vx,\vtheta)$ misses on underrepresented instances.}
	\label{fig:prob}
\end{figure}

The existing AL methods assume that distribution of collected train examples is somewhat similar to test cases and, hence, relevant data points can be found only by accessing train data. This assumption rarely holds for the unlabeled data where very \textit{rare examples} have to be identified as illustrated in Figure~\ref{fig:prob}. The classifier learned on train data selected by the existing AL methods can have high error rate on underrepresented instances. For example, distribution of digits "1" prevails over \textit{rare} digits "9" in train data and, as a result, test digits "9" are misclassified. Moreover, consider an autonomous vehicle only trained to perform well in the most frequent conditions rather than in a rare critical situations such as car crashes. To overcome this limitation, we propose a new acquisition function for AL. It is based on distribution matching between the validation dataset and the AL-selected training data. Validation dataset in such setting covers important cases from the long-tail of distribution that can be continuously identified and added after field trials.

We achieve distribution matching by pooling multi-scale low-dimensional discriminative features from the \textit{task classifier model}. Our key contribution is the usage of Fisher kernel (FK) to find the most important examples with the improved pseudo-label estimators using several novel metrics. Finally, we incorporate recent unsupervised pretraining method~\cite{gidaris} to speed up representation learning by the task model. Our framework is well-suited for large-scale data because its complexity is only a single forward and backward pass per data point. We show the effectiveness of our method on MNIST, SVHN, and ImageNet classification including biased training data with long-tailed distribution, where the proposed method is able to decrease labeling efforts by at least 40\% compared to prior methods.

\section{Related work}
\label{sec:related}
AL is a well-studied approach to decrease annotation efforts in traditional machine learning pipelines~\cite{settles}. Recently, AL has been applied to DNN-based models in semi-supervised setting with oracle labeling or weakly-supervised setting with pseudo-labeling. While our method can be applied to both types, we mainly focus on prior work of a more robust semi-supervised pool-based AL.

Gal~\etal~\cite{gal} introduced a measure of uncertainty for approximate Bayesian inference that can be estimated using stochastic forward passes through a DNN with dropout layers. Their AL acquisition function selects data points with the highest uncertainty which is measured at the output of classifier's \textit{softmax} layer using several metrics. Recent work by Beluch~\etal~\cite{bel} extended this method by using an ensemble of networks for uncertainty estimation and achieved superior accuracy.

Sener and Savarese~\cite{sener} formulated training dataset selection for AL as a geometric \textit{core-set} clustering approach which outperforms greedy $k$-center clustering. Though their core-set clustering can complement our approach, we are focusing on a discriminative low-dimensional feature extraction followed by inexpensive clustering. Computational complexity of the core-set clustering is a potential bottleneck where two orders of magnitude more processing is needed compared to greedy clustering in our approach. 

Recently, Sinha~\etal~\cite{vaal} proposed to use variational autoencoder (VAE)~\cite{vae} to learn a latent space followed by an adversarial network~\cite{makhzani} to discriminate between labeled and unlabeled data. Their AL acquisition function is the output of discriminator, which \textit{implicitly} learns the most likely to be labeled examples. This variational adversarial active learning (VAAL) approach claims to achieve superior results compared to all previous works. However, VAAL has large number of hyperparameters and high complexity since VAE and discriminator have to be retrained on all unlabeled and labeled train data every AL iteration.

The closest to our method, line of works~\cite{if,khanna} employs influence functions and Fisher kernels as a measure of feature importance for dataset subsampling and analysis. Khanna~\etal~\cite{khanna} showed equivalence of FK and influence functions for $\log$-likelihood loss functions. Similar work on online importance sampling using Fisher score similarity~\cite{l2r} upweights samples within the mini-batch during fully-supervised training. However, these approaches require fully-labeled data to estimate FK.

Another related area is unsupervised representation learning that, unfortunately, has not been used in AL literature. At the same time, recent approaches~\cite{gidaris,cpc,infomax} significantly improved previous state-of-the-art. Hence, we incorporate unsupervised pretraining into our AL method to speed up latent representation learning.

The existing methods struggle to deal with biased data as sketched in Figure~\ref{fig:prob}. Motivated by this, we develop our framework with the following contributions:
\begin{itemize}
	\itemsep0em
	\item We derive an optimal acquisition function $\mathcal{R}_{opt}(\cdot)$ for biased datasets, which is formulated as a task to minimize Kullback–Leibler (KL) divergence between distributions of training and validation datasets.
	\item We propose a low-complexity non-parametric AL method via self-supervised FK using a set of pseudo-label estimators and derive its connection to $\mathcal{R}_{opt}(\cdot)$. 
	\item We complement our method by the recent unsupervised pretraining method using image rotations~\cite{gidaris}.
	\item Our method outperforms prior methods in image classification. In particular, datasets with long-tailed biased train data result in at least 40\% less labeling.
\end{itemize}

\section{Problem statement for biased datasets}
\label{sec:theory}
Let $(\vx,\vy)$ be an input-label pair where a label $\vy$ is one-hot vector for a classification task. There is a relatively small validation dataset $\valid=\{(\vx^\mathrm{v}_i,\vy^\mathrm{v}_i)\}_{i\in{\sM}}$ of size $M$ and a large collection of training pairs $\train=\{(\vx_i,\vy_i)\}_{i\in{\sN}}$ of size $N$ for which, initially, all labels are unknown. The validation dataset can be weakly labeled as discussed below. At every $b$th iteration AL acquisition function $\mathcal{R}(\cdot)$ selects a pool of $P$ new labels to be annotated and added to train data which creates a training dataset indexed by subset $\sN^b$.

A feed-forward DNN model $f(\vx,\vtheta)$ is optimized with respect to parameter vector $\vtheta$ using supervised learning framework by minimizing objective function
\begin{equation} \label{eq:t1}
\mathcal{L}(\vtheta) = \frac{1}{N^b} \sum_{i \in \sN^b} L(\vy_i,\hat{\vy}_i) = \frac{1}{N^b} \sum_{i \in \sN^b} L(\vy_i,f(\vx_i,\vtheta)),
\end{equation}
where $L(\vy_i, \hat{\vy}_i)$ is a loss function and $\hat{\vy}_i$ is output prediction. The loss function is a negative $\log$ probability of discrete $\vy$ for classification task. This is equivalent to minimization of approximate KL divergence $D_{KL}$ between joint training data distribution $Q_{\vx,\vy}$ with density $q(\vx,\vy)$ and the learned model distribution $P_{\vx,\vy}(\vtheta)$ with corresponding density $p(\vx,\vy | \vtheta)$. Since $q(\vx,\vy) = q(\vy | \vx) q(\vx)$ and $p(\vx,\vy | \vtheta) = p(\vy | \vx, \vtheta) q(\vx)$, KL objective learns only \textit{conditional distribution} of $\vy$ given $\vx$ as
\begin{equation} \label{eq:t2}
\begin{split}
D_{KL}(Q_{\vx,\vy} \| P_{\vx,\vy}(\vtheta)) &= \\
\int q(\vx) \int q(\vy | \vx) & \log \frac{q(\vy | \vx) q(\vx)}{p(\vy | \vx, \vtheta) q(\vx)} d\vy d\vx = \\
&\sE_{Q_\vx}[ D_{KL}(Q_{\vy|\vx} \| P_{\vy|\vx}(\vtheta) ) ].
\end{split}
\end{equation}

Due to unknown density $q(\vx)$, the expectation over $Q_{\vx}$ in (\ref{eq:t2}) is usually replaced by \textit{empirical distribution} $\hat{Q}_{\vx}$ as
\begin{equation} \label{eq:t3}
\begin{split}
\sE_{\hat{Q}_\vx} & [ D_{KL}(Q_{\vy|\vx} \| P_{\vy|\vx}(\vtheta) ) ] = \\
& \frac{1}{|\train|} \sum_{(\vx,\vy) \in \train} [ D_{KL}(Q_{\vy|\vx} \| P_{\vy|\vx}(\vtheta) ) ].
\end{split}
\end{equation}

By rewriting loss $L(\cdot)$ in (\ref{eq:t1}) using $D_{KL}$ from (\ref{eq:t3}), objective function $\mathcal{L}(\vtheta)$ can be rewritten as negative $\log$ of conditional probability
\begin{equation} \label{eq:t4}
\mathcal{L}(\vtheta) = -\frac{1}{N^b} \sum\nolimits_{i \in \sN^b} \log p(\vy_i|\vx_i,\vtheta).
\end{equation}

However, the \textit{actual task} is to minimize objective~(\ref{eq:t2}) for test data $\test$ with expectation over $Q_{\vx}^{\mathrm{test}}$ distribution. This contradiction is usually resolved in AL literature by assuming $Q_{\vx}^{\mathrm{test}}$ and $Q_{\vx}$ equality. In practice, the deployed systems struggle to deal with underrepresented test cases in the train distribution $Q_{\vx}$. The examples include autonomous vehicles in rare traffic situations or facial recognition systems with gender and race biases~\cite{shades}. This is schematically illustrated in Figure~\ref{fig:prob}.

We argue that the \textit{key requirement} for effective AL \textit{in the wild} is to collect a validation dataset $\valid$ with distribution $Q_{\vx}^{\mathrm{v}}$, which approximates $Q_{\vx}^{\mathrm{test}}$. To be specific, we approximate distribution of a representative collection of \textit{test cases} in $\valid$ and continuously update it by newly discovered misclassified data. This can be done iteratively after conducting field trials for deployed systems. The assumptions about $\valid$ and $\train$ are summarized in Figure~\ref{fig:0}.

It follows from~(\ref{eq:t2}) that an optimal acquisition function $\mathcal{R}_{opt}(\cdot)$ for AL minimizes distribution shift between $\test$ and $\train$, where the former is approximated by empirical $\valid$. This can be expressed using KL divergence as
\begin{equation} \label{eq:t5}
\begin{split}
\mathcal{R}_{opt}(b,P) = \argmin_{\mathcal{R}(b,P)} D_{KL}&(Q_{\vx}^{\mathrm{test}} \| Q_{\vx}) \approx \\
& \argmin_{\mathcal{R}(b,P)} D_{KL}(\hat{Q}_{\vx}^{\mathrm{v}} \| \hat{Q}_{\vx}),
\end{split}
\end{equation}
where, in practice, (\ref{eq:t5}) can be replaced by locally optimal steps for every iteration $b=1 \ldots B$ and pool size $P$.

\begin{figure}[t]
	\centering
	\includegraphics[width=0.98\columnwidth]{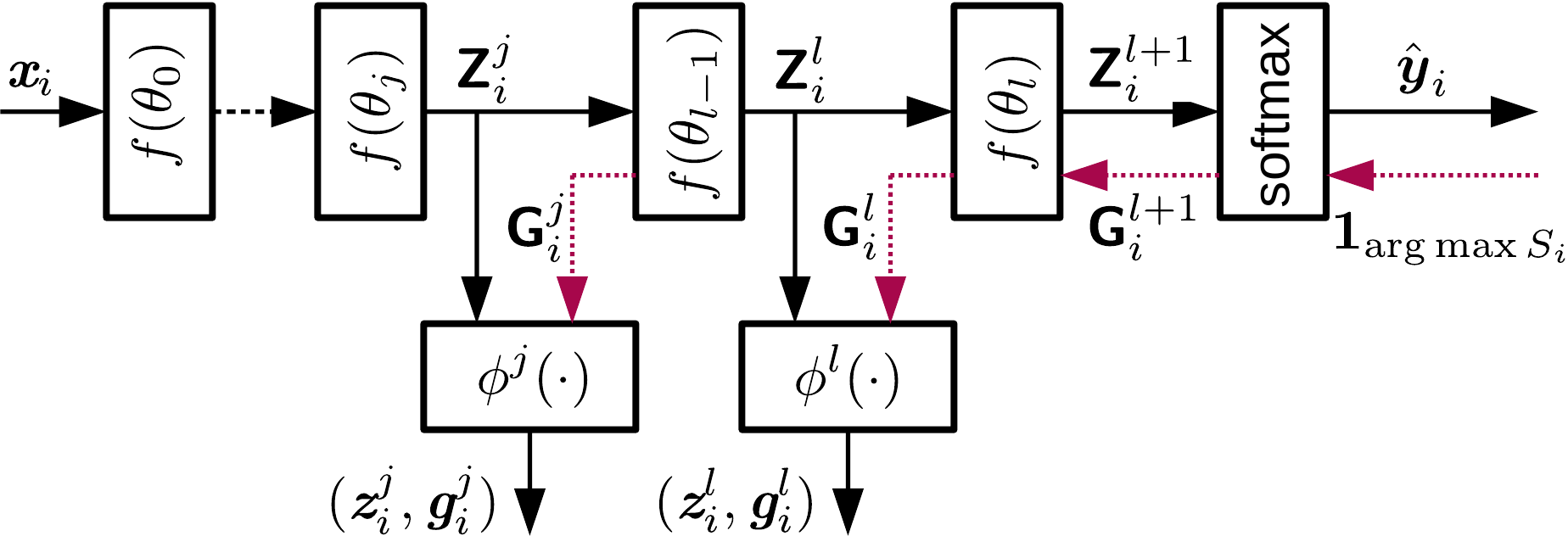}
	\caption{Conventional multi-scale feature extraction and the proposed FK extension (\textcolor{purple}{dashed}). Descriptors $\vz_i$ and Fisher score vectors $\vg_i$ are used for density matching by our AL method.}
	\label{fig:1}
\end{figure}

\section{The proposed method}
\label{sec:method}

\subsection{Conventional feature descriptors for AL} \label{subsec:prop-feat}
High dimensionality of input $\vx$ causes computational difficulties in minimizing (\ref{eq:t5}). Then, $\vx$ is usually replaced by a low-dimensional feature \textit{descriptor} in image retrieval~\cite{zheng}. Such descriptors are pooled from DNN intermediate representations $\vz$, which are found to be effective~\cite{babenko}. Then,~(\ref{eq:t5}) can be reformulated as empirical distribution matching between $\hat{P}_{\vz}^{\mathrm{v}}$ and $\hat{P}_{\vz}$. This can be done using various methods~\cite{gretton}, but, practically, a greedy $k$-center clustering for density estimation with a similarity measure is the most used method for the large train dataset size $N$.

Let $\tZ^j_i \in\mathbb{R}^{C \times H \times W}$ be the output of $j$th layer of \textit{task DNN model} for input image $\vx_i$ as shown in Figure~\ref{fig:1} for image classification, where $C$, $H$, and $W$ are the number of channels, the height, and the width, respectively. Then, a feature vector or descriptor of length $L$ can be defined as $\vz_i=\phi(\tZ_i)\in\mathbb{R}^{L}$, where function $\phi(\cdot)$ is a conventional average pooling operation. In a multi-scale case, descriptor $\vz_i$ is a list of multiple feature vectors $\vz^j_i$.

A descriptor matrix for the validation dataset $\mZ_\mathrm{v}\in\mathbb{R}^{L \times M}$ and training dataset $\mZ\in\mathbb{R}^{L \times N}$ can be efficiently calculated using DNN forward passes. Practically, descriptors can be further compressed for storage efficiency reasons using PCA, quantization, etc. Pearson correlation (PCC) is a common match kernel, which is an accurate measure of \textit{linear} correlation. By preprocessing vectors $\vz_i$ to have zero mean and unit variance, the similarity (cross-covariance) matrix for multi-scale case is simply
\begin{equation} \label{eq:p2}
\mR_\vz = \sum\nolimits_{j} (\mZ_\mathrm{v}^j)^T \mZ^j.
\end{equation}

Using information theory~\cite{gray}, this framework assumes representation $\vz$ to have the following properties about the task: \textit{minimality} $(\min I(\vz;\vx))$ and \textit{sufficiency} $(I(\vy;\vz)=I(\vy;\vx))$, where $I()$ is mutual information quantity. Indeed, Achille~\etal~\cite{achille} analytically shows that a DNN trained by stochastic gradient descent (SGD) discards non-informative features and retains only the ones to minimize objective function~(\ref{eq:t2}). However, these properties are applicable only for a fully trained model without bias in train data.

An alternative approach is to use an autoencoder~\cite{NIPS1993_798} or, similarly to VAAL~\cite{vaal}, probabilistic VAE~\cite{vae} to compress $\vx$ to $\vz$. Those alternatives require to train another model using a new set of hyperparameters and reconstruction loss rather than task-specific objective~(\ref{eq:t2}). However, the learned representation $\vz$ is subject to biased train data pitfall shown in Figure~\ref{fig:prob}. Fortunately, this pitfall can be resolved for \textit{task model} by AL itself, if it minimizes distribution shift in~(\ref{eq:t5}). Hence, we choose to pool features $\vz$ from the \textit{task model} in our framework to avoid data bias, additional complexity, and hyperparameter search issues. We address the sufficiency property discussed above by using unsupervised pretraining followed by a more powerful match kernel.

\subsection{Self-supervised Fisher kernel} \label{subsec:prop-fisher}
Recent works~\cite{if,khanna} revived interest in influence functions and Fisher kernels used in pre-DNN era~\cite{old-fisher}. They are able to identify the most influential training points for a given test data. Though attractive, these methods are computationally expensive for large-scale data and DNN models because FK is typically calculated with respect to high-dimensional parameter vector $\vtheta$.

Using the sufficiency property~\cite{achille}, we approximate our optimal acquisition function~(\ref{eq:t5}) using the distributions of learned representations $\vz$ as
\begin{equation} \label{eq:p22}
\mathcal{R}_{opt}(b,P) = \argmin_{\mathcal{R}(b,P)} D_{KL}(\hat{P}_{\vz}^{\mathrm{v}} \| \hat{P}_{\vz}),
\end{equation}

Then, a \textit{connection} between the main task~(\ref{eq:t2}) and $D_{KL}(P_{\vz}^{\mathrm{v}} \| P_{\vz})$ minimization in~(\ref{eq:p22}) via Fisher information can be derived with respect to small perturbations in $\vtheta$. Assuming that the task model minimizes distribution shift in~(\ref{eq:t2}) every backward pass as
\begin{equation} \label{eq:p23}
p^{\mathrm{v}}(\vz|\vtheta) = p(\vz|\vtheta) + \Delta p,
\end{equation}
where $\Delta p = \Delta \vtheta \partial p(\vz|\vtheta) / \partial \vtheta$ and $\Delta \rightarrow 0$.

By substituting~(\ref{eq:p23}), the expanded form of (\ref{eq:p22}) can be simplified using Taylor series of natural logarithm as
\begin{equation} \label{eq:p25}
\mathcal{R}_{opt}(b,P) \approx \argmin_{\mathcal{R}(b,P)} \Delta \vtheta^T \mathcal{\mI} \Delta \vtheta,
\end{equation}
where $\mathcal{\mI} = \sE_{P_\vz} \left[ \vg(\vtheta) \vg(\vtheta)^T \right]$ is a Fisher information matrix and $\vg(\vtheta) = \partial \log p(\vz|\vtheta)/\partial \vtheta$ is a Fisher score with respect to $\vtheta$. The detailed derivation is given in Appendix.

Using result in (\ref{eq:p25}), Jaakkola and Hauusler~\cite{jakkola} proposed the popular Fisher kernel expressed by
\begin{equation} \label{eq:p3}
\emR_{z,g}(\vz_m, \vz_n) = \vg_m(\vtheta)^T \mathcal{\mI}^{-1} \vg_n(\vtheta).
\end{equation}

To make~(\ref{eq:p3}) computationally tractable, we use \textit{practical} FK (PFK) where $\mathcal{\mI}^{-1}$ is replaced by identity matrix. Such a common approach decreases quadratic storage requirements. Next, we rewrite Fisher scores $\vg_i(\vtheta)$ using a more compact form $\vg_i(\vtheta) = \mathrm{vec}(\vg_i \vz^T_i )$, where $\vg_i$ is computed with respect to features as $\vg_i = \partial  L(\vy_i, \hat{\vy}_i) / \partial \tilde{\vz}_i $, $L(\vy_i, \hat{\vy}_i)$ is $\log$-likelihood loss function from (\ref{eq:t4}), and $\tilde{\vz}_i$ is a vector before applying nonlinearity $\sigma(\cdot)$. The latter follows from the chain rule when computing loss function for a DNN layers $(\tilde{\vz}^j_i = \vtheta^T \vz^j_i = \vtheta^T \sigma(\tilde{\vz}^{j-1}_i))$ as derived in Appendix. Then, the tractable PFK can be rewritten for DNNs as
\begin{equation} \label{eq:p32}
\emR_{z,g}(\vz_m, \vz_n) = \vg_m(\vtheta)^T \vg_n(\vtheta) = \vz^T_m \vz_n \vg^T_m \vg_n.
\end{equation}

Fisher scores in (\ref{eq:p32}) are also related to visual explanation methods~\cite{montavon}. If replace $\vz$ with $\vx$ in $\vg(\vtheta)$ calculation, the result estimates popular importance heatmaps in the input space. In our case, kernel~(\ref{eq:p32}) shows the model sensitivity to changes in parameters caused by distribution shift $D_{KL}(\hat{P}_{\vz}^{\mathrm{v}} \| \hat{P}_{\vz})$. Then, PFK matrix $\mR_{\vz,\vg}\in\mathbb{R}^{M \times N}$ can be efficiently calculated using a series of forward-backward passes. By analogy to feature similarity (\ref{eq:p2}), the Fisher scores $\vg_i$ for images are calculated with respect to tensors $\tZ_i$ and pooled by the same $\phi(\cdot)$ such that $\vg_i = \phi(\partial L_i / \partial \tZ_i)\in\mathbb{R}^{L}$. Finally, we minimize the distribution shift in~(\ref{eq:p22}) by maximizing PFK as
\begin{equation} \label{eq:p4}
\mathcal{R}_{opt}(b,P) = \argmax_{\mathcal{R}(b,P)} \mR_{\vz,\vg},
\end{equation}
where $\mR_{\vz,\vg} = \mR_\vz \circ \mR_\vg = \sum\nolimits_{j} (\mZ_\mathrm{v}^j)^T \mZ^j \circ (\mG_\mathrm{v}^j )^T \mG^j$. Our PFK matrix $\mR_{\vz,\vg}$ is an element-wise multiplication of feature similarity from~(\ref{eq:p2}) and gradient similarity matrices.

\subsection{The proposed pseudo-label estimators} \label{subsec:prop-pseudo}
The main drawback of (\ref{eq:p4}) is lack of labels $\vy$ in the unlabeled collection of train data. The common pseudo-labeling $(\vone_{\argmax_d S})$ metric $S(\cdot)$ assigns hard-label to the $d$th class with maximum predicted probability: $S=\hat{\vy}$. That leads to incorrect estimates during first AL iterations, particularly, for rare examples. To overcome this limitation, we propose several novel metrics to estimate pseudo-labels.

First, we introduce estimation metrics using Monte Carlo (MC) sampling. Consider a DNN input $\vx$  being sampled near its local neighborhood. That produces inputs $\vx_k$, feature samples $\vz_k$, and a corresponding per-class Fisher scores $\vg_k(d) = \partial L(\vone_{d}, \hat{\vy}_k) / \partial \vz_k$, where class $d=1 \ldots D$. The sampling can include small rotations, translations or color distortions for image inputs~\cite{amdim}. The simplest MC label estimation maximizes linear correlation between features and Fisher scores as $S=\tr{(\mC_{\vz,\vg})}$, where $\mC_{\vz,\vg}$ is cross-covariance matrix between feature descriptors and Fisher scores. Theoretically, a better metric is maximization of mutual information $I(\vz; \vg)$ to capture nonlinear dependency. Classic result~\cite{gray} shows that for random vectors $\vz$ and $\vg$ that follow Gaussian probability model, average mutual information can be estimated as $S=I(\vz; \vg) = 0.5 \log \left( \left|\mC_{\vz,\vz}\right| \left|\mC_{\vg,\vg}\right| / \left|\mC_{\vz\vg,\vz\vg}\right| \right)$, where $\left|\mC\right|$ is the determinant of cross-covariance matrix. This can be efficiently calculated using LU or Cholesky decomposition implemented in modern ML frameworks~\cite{paszke2017automatic}.

The second proposed metric explicitly estimates $\hat{p}(\vy,\vz) = \hat{p}(\vy|\vz) p(\vz)$, for which it is necessary to have a trusted annotated dataset to obtain $\hat{p}(\vy|\vz)$. In our case, it can be validation dataset $\valid$ or its subset. Since $p(\vy|\vz) = p^\mathrm{v}(\vy|\vz)$, the estimate $\hat{p}(\vy,\vz)$ can be found from trusted conditional density $p^\mathrm{v}(\vy|\vz)$ and marginal $p(\vz)$. We propose to reuse the described above framework to find the most similar data points in $\valid$ to examples in $\train$ using $\mR_{\vz}$ kernel. Then, we assign given trusted labels $\vy^\mathrm{v}$ from $p^\mathrm{v}(\vz)$ to train labels from $p(\vz)$ for which $\mR_{\vz}$ is maximized. This results in a low-complexity non-parametric method.

To summarize, we experiment with the following label estimation metrics: a) $S=\vy$ for ablation study with true labels, b) common $S=\hat{\vy}$, as well as the proposed c) MC $S=\tr{(\mC_{\vz,\vg})}$, d) MC $S=I(\vz;\vg)$ and e) $S=\hat{p}(\vy,\vz)$.

\subsection{Complexity of weakly-supervised algorithm} \label{subsec:prop-sum}
While FK finds the most similar data points using discriminative representation, our AL needs to identify validation points for distribution matching using~(\ref{eq:p4}). However, even inexpensive greedy $k$-center clustering might be prohibitive $(\mathcal{O}(PM))$ for relatively small $\valid$. To address this, we propose to use \textit{weak supervision} (correct or incorrect prediction) to find subset of misclassified validation examples $\{\vone_{\argmax_d \hat{\vy}^\mathrm{v}_i} \neq \vy^\mathrm{v}_i\}_{i \in \acute{\sM}}$, where $\acute{M} < M$. Then, this subset is clustered using $k$-centers, and $P$ validation points are selected to maximize PFK in~(\ref{eq:p4}). Weak supervision assumption typically holds because, often, $\valid$ is already fully-labeled to know how model is performing. Variant of our weakly-supervised method is fully described in Alg.~\ref{alg:1}.

Computational complexity of PFK is estimated in Table~\ref{tab:1} in terms of forward and backward DNN passes. Note that the complexity of greedy clustering, finding cross-covariance matrices is not shown because it is negligible compared to DNN passes. For comparison with AL phase (lines 3-10 in Alg.~\ref{alg:1}), we report complexity of retraining phase (line 11) using $I$ epochs and $N^b$ labeled train data.

Since the number of unlabeled data $\acute{N}^b$ ($\acute{N}^b = N - N^{b-1}$ in line 8) is much bigger than validation data $M$, our method is $EK/2$ times less complex than uncertainty methods~\cite{gal,bel} with $K$ stochastic passes and $E$ ensembles.

VAAL~\cite{vaal} consists of sampling phase with $\acute{N}^b$ forward passes and retraining phase of VAE and discriminator models using $I_{\textrm{VAE,D}}$ epochs. Assuming that VAE, discriminator and task model $f(\vx,\vtheta)$ have roughly the same complexity, our method is $I_{\textrm{VAE,D}}$ times less complex than VAAL.

The method with PCC kernel~(\ref{eq:p2}) is $2\times$ less complex than ours with PFK. The variant of our method with MC pseudo-labeling $(S=\tr{(\mC_{\vz,\vg})}~\textrm{or}~I(\vz;\vg))$ is $KD/2$ times more complex than PFK with inexpensive metrics $(S=\hat{\vy}~\textrm{or}~\hat{p}(\vy,\vz))$, where $D$ is number of classes. MC metrics have potentially better accuracy compared to $S=\hat{\vy}$ without reliance on a trusted labeled dataset as in $S=\hat{p}(\vy,\vz)$.

\begin{algorithm}[t]
	\caption{Variant with weakly-supervised $\valid$.}
	\label{alg:1}
	\begin{algorithmic}[1]
		\State \textbf{Initialize:} $\sN^{0} = \{\}$, $\vtheta^0$ random or pretrained by~\cite{gidaris}
		\For{$b=1, 2 \ldots B$}
		\State find misclassified subset $\{\vone_{\argmax_d \hat{\vy}^\mathrm{v}_i} \neq \vy^\mathrm{v}_i\}_{i \in \acute{\sM}}$
		\State pool matrices $(\mZ_\mathrm{v},\mG_\mathrm{v})\in\mathbb{R}^{L \times \acute{M}}$
		\If{$\acute{M} > P$}
		\State find $P$ centers in $\acute{\sM}$ using $k$-center clustering
		\State subsample matrices $(\mZ_\mathrm{v}, \mG_\mathrm{v})\in\mathbb{R}^{L \times P}$
		\EndIf
		\State pool matrices $(\mZ,\mG)\in\mathbb{R}^{L \times \acute{N}^b}$, $\acute{N}^b=N - N^{b-1}$
		\State calculate PFK matrix $\mR_{\vz,\vg} = \mR_\vz \circ \mR_\vg$
		\State add $P$ points to $\sN^b$ as $\argmax_p \mR_{\vz,\vg}$
		\State update $\vtheta^b = \argmin_\vtheta \sum\nolimits_{i \in \sN^b} L(\vy_i, \hat{\vy}_i) / N^b$
		\EndFor
	\end{algorithmic}
\end{algorithm}

\begin{table}
	\caption{Complexity estimates per AL iteration. Assuming $\acute{N}^b>>M$, our method has the lowest complexity in terms of forward and backward DNN passes during AL phase.}
	\label{tab:1}
	\centering
	\begin{tabular}{lcc}
		\toprule
		Method & AL & Train\\
		\midrule
		Uncert.~\cite{gal}                   & $K\acute{N}^b$                       & $2IN^b$\\
		Ens. uncert.~\cite{bel}              &$EK\acute{N}^b$                       &$2EIN^b$\\
		VAAL~\cite{vaal}                     &$\acute{N}^b+2NI_{\textrm{VAE,D}}$      & $2IN^b$\\
		PCC~(\ref{eq:p2}): $\mR_\vz$         &$  M+\acute{N}^b $                    & $2IN^b$\\
		PFK~(\ref{eq:p4}): $\mR_{\vz,\vg}$~(\textbf{ours})    &$2(M+\acute{N}^b)$   & $2IN^b$\\
		PFK\textsubscript{MC}(\ref{eq:p4}): $\mR_{\vz,\vg}$~(\textbf{ours})         &$KD(M+\acute{N}^b)$ & $2IN^b$\\
		\bottomrule
	\end{tabular}
\end{table}

\begin{figure*}[t]
\centering
    \begin{tabular}{ccc}
        \includegraphics[width=0.32\textwidth]{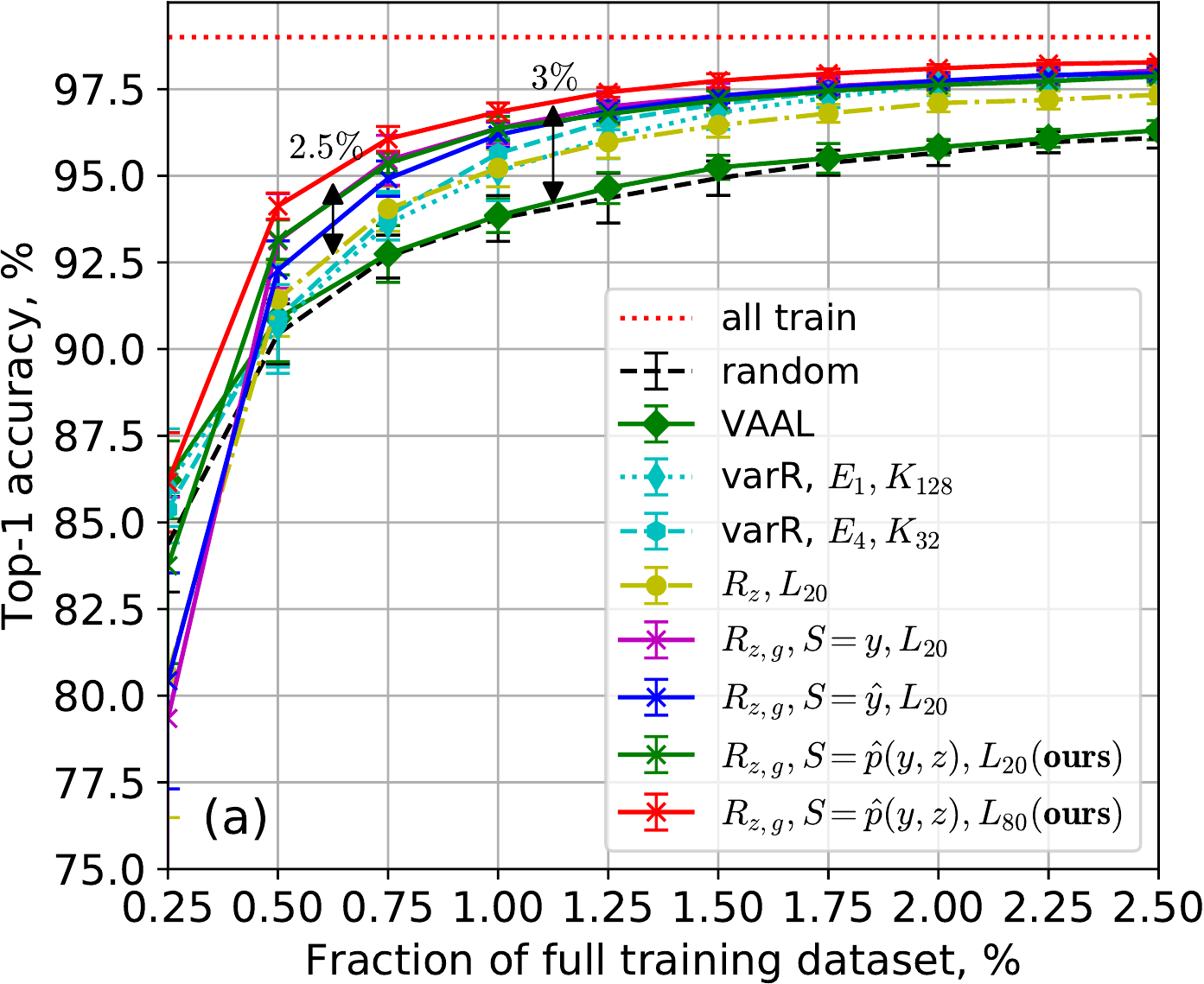} &
        \includegraphics[width=0.32\textwidth]{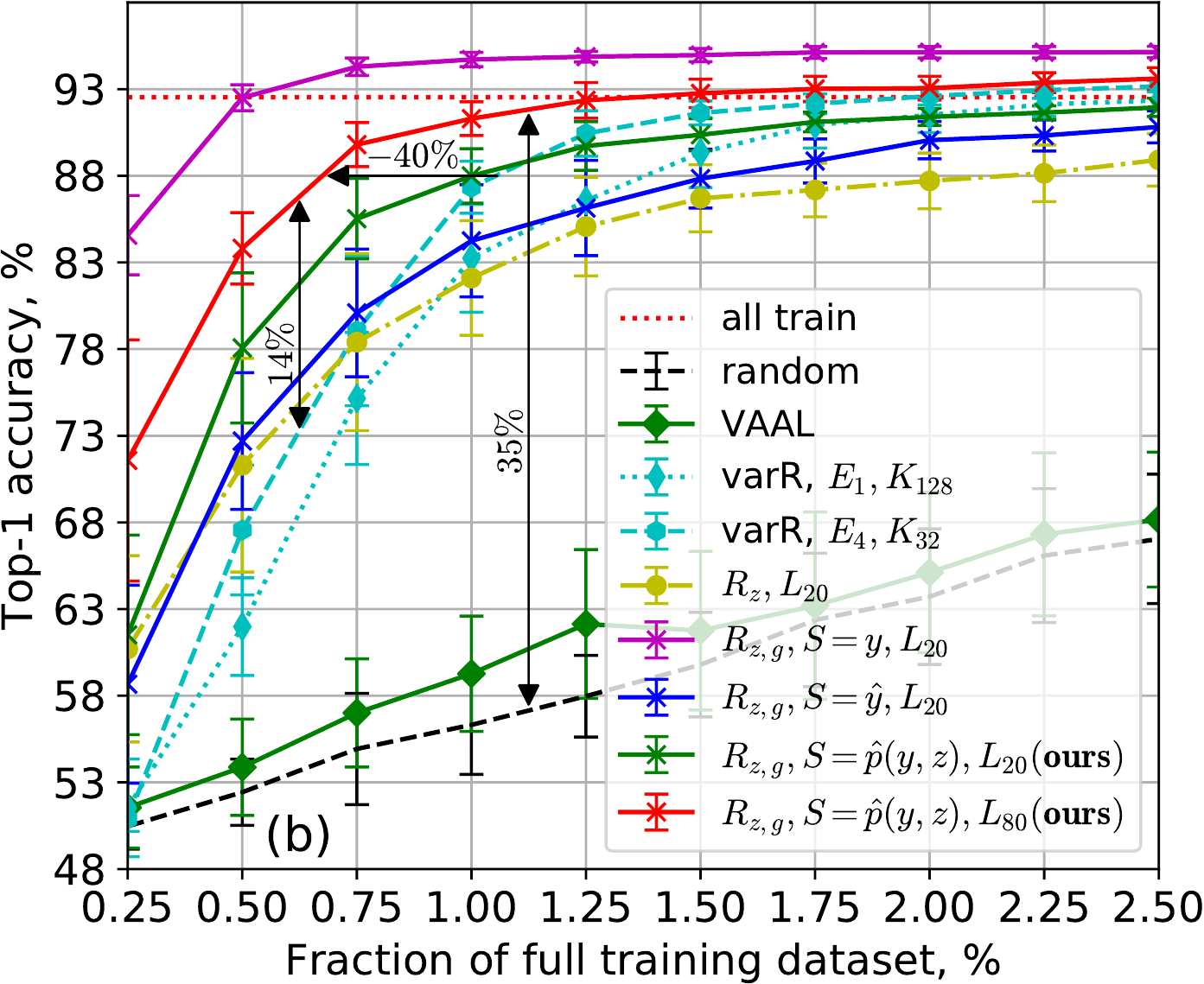} &
        \includegraphics[width=0.32\textwidth]{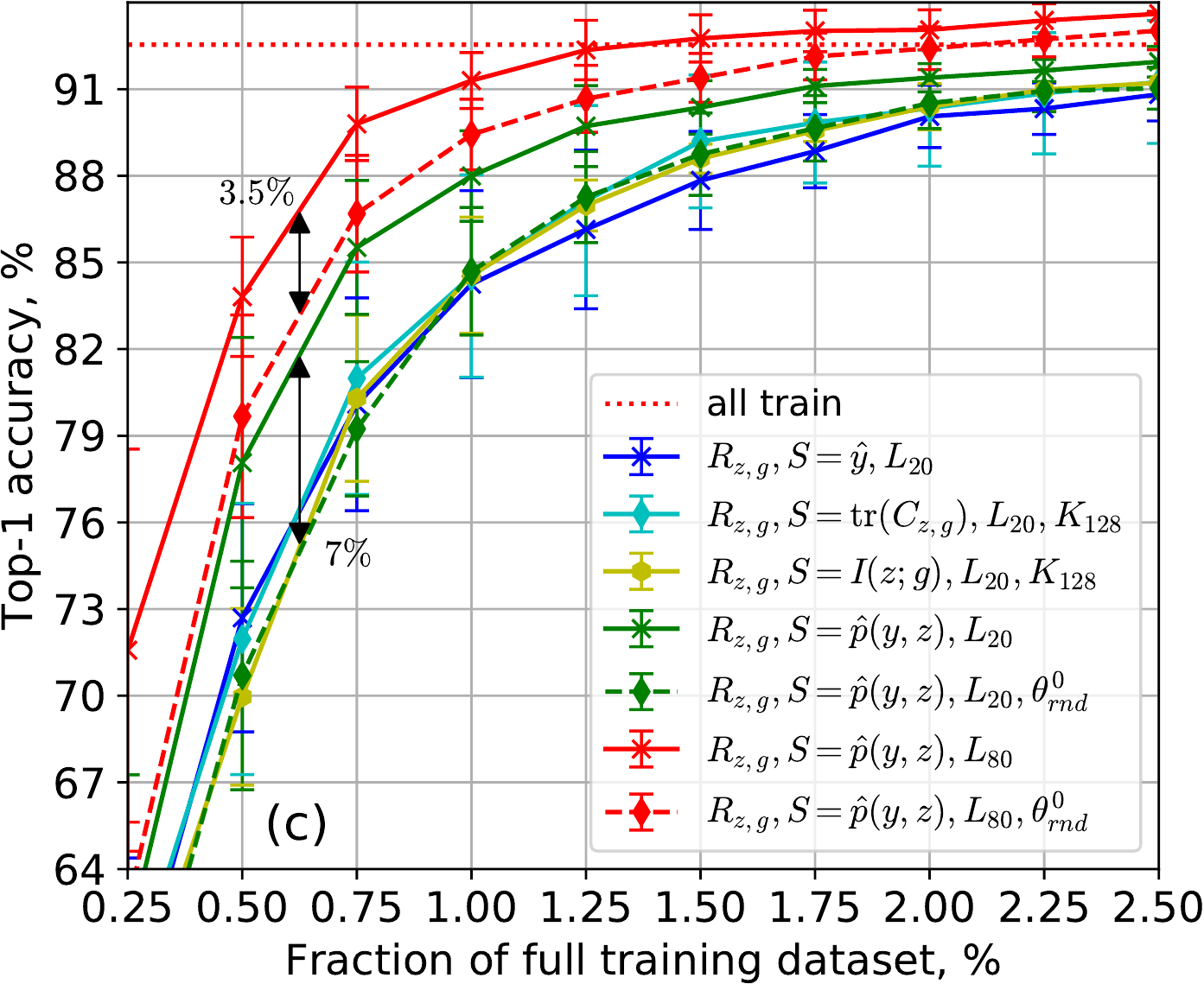}
    \end{tabular}
    \caption{MNIST test accuracy: (a) no class imbalance, (b) $100\times$ class imbalance, and (c) ablation study of pseudo-labeling and unsupervised pretraining ($100\times$ class imbalance). Our method decreases labeling by 40\% compared to prior works for biased data.}
    \label{fig:mnist-fig}
\end{figure*}

\section{Experiments}
\label{sec:evaluation}
We apply our framework to MNIST, SVHN and ImageNet classification. We evaluate AL not only with the original training data, but also their biased versions. Hence, we introduce a \textit{class imbalance} which scales down number of available train images for subset of classes. Class imbalance is defined as the ratio of $\{0\ldots4\}$ digits to $\{5\ldots9\}$ digits for MNIST and SVHN. We randomly select 500 out of 1,000 classes for ImageNet. Train examples for the selected 500 classes are decimated by the class imbalance ratio, while the other 500 classes keep the original train data. The code is written in PyTorch~\cite{paszke2017automatic} with reproducible experiments and is publicly available.

The following experimental configurations are defined: baseline when \textit{all train} data is used, \textit{random} sampling, and methods from Table~\ref{tab:1}. We reimplemented all uncertainty methods~\cite{gal,bel}: variation ratio (\textit{varR}), maximum entropy and BALD. Only results of the best-performing varR method are reported. We use official code for VAAL~\cite{vaal} experiments. We use the following notation in figures: number of ensembles is specified by the $E$, samples by $K$, and descriptor size by $L$. 

We run each experiment $10\times$ for MNIST, $5\times$ for SVHN and once for large-scale ImageNet on V100 GPUs. We report mean accuracy and standard deviation for MNIST and SVHN test dataset. Due to lack of test labels for ImageNet, we use validation dataset for testing. Each AL experiment consists of 10 iterations $(B=10)$. With the exception of last fully-connected layer, initial network parameters are from unsupervised pretraining using rotation method~\cite{gidaris} or, if specified, randomly initialized. Large batch sizes may underperform with class-imbalanced data and, therefore, we select mini-batch size by cross-validation. The used DNN models are LeNet, ResNet-10, and ResNet-18 for MNIST, SVHN, and ImageNet, respectively. The dropout configurations are the same or similar to~\cite{gal,bel} setups.

\begin{figure*}[t]
	\centering
	\begin{tabular}{cc}
		\includegraphics[width=0.32\textwidth]{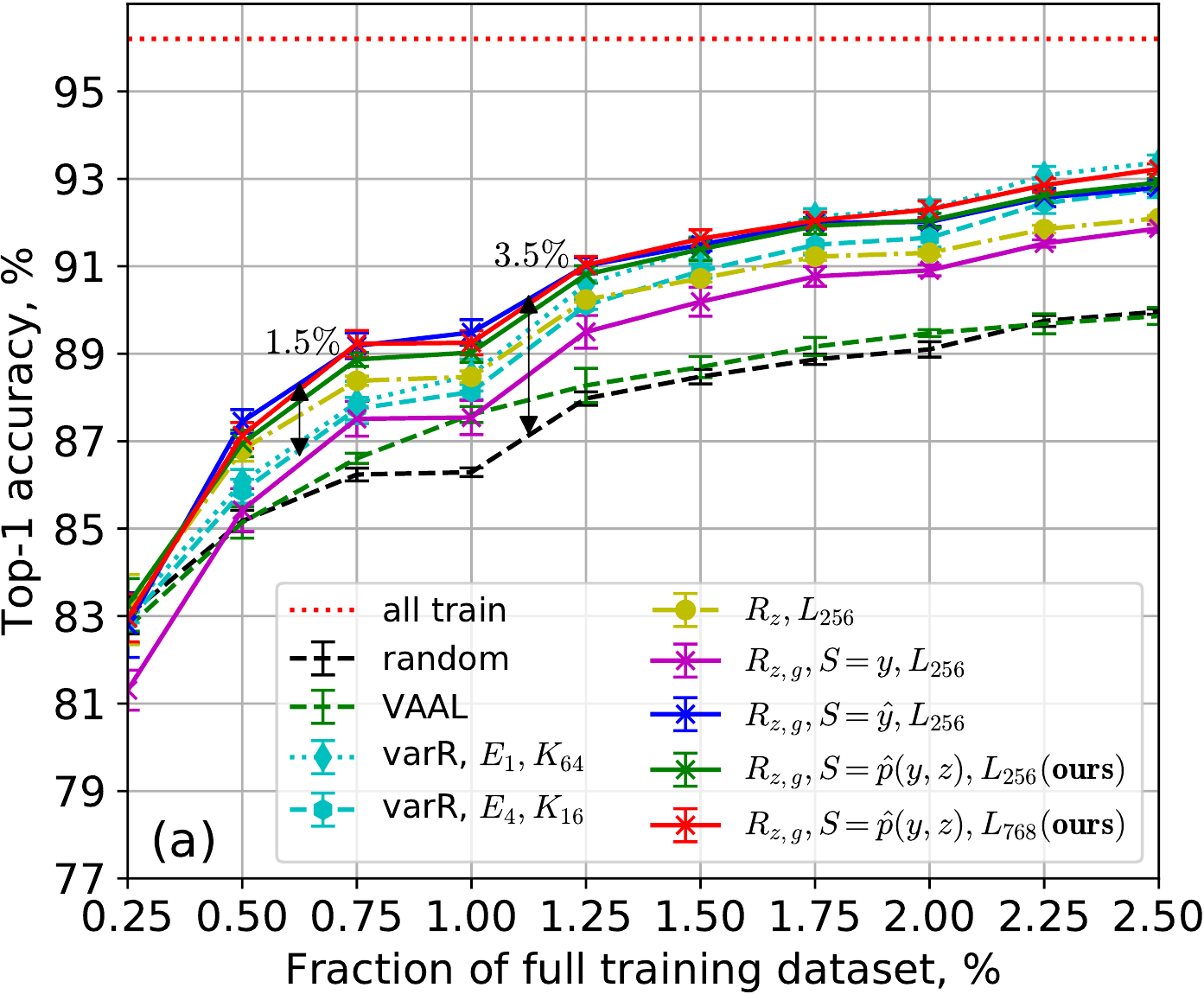} &
		\includegraphics[width=0.32\textwidth]{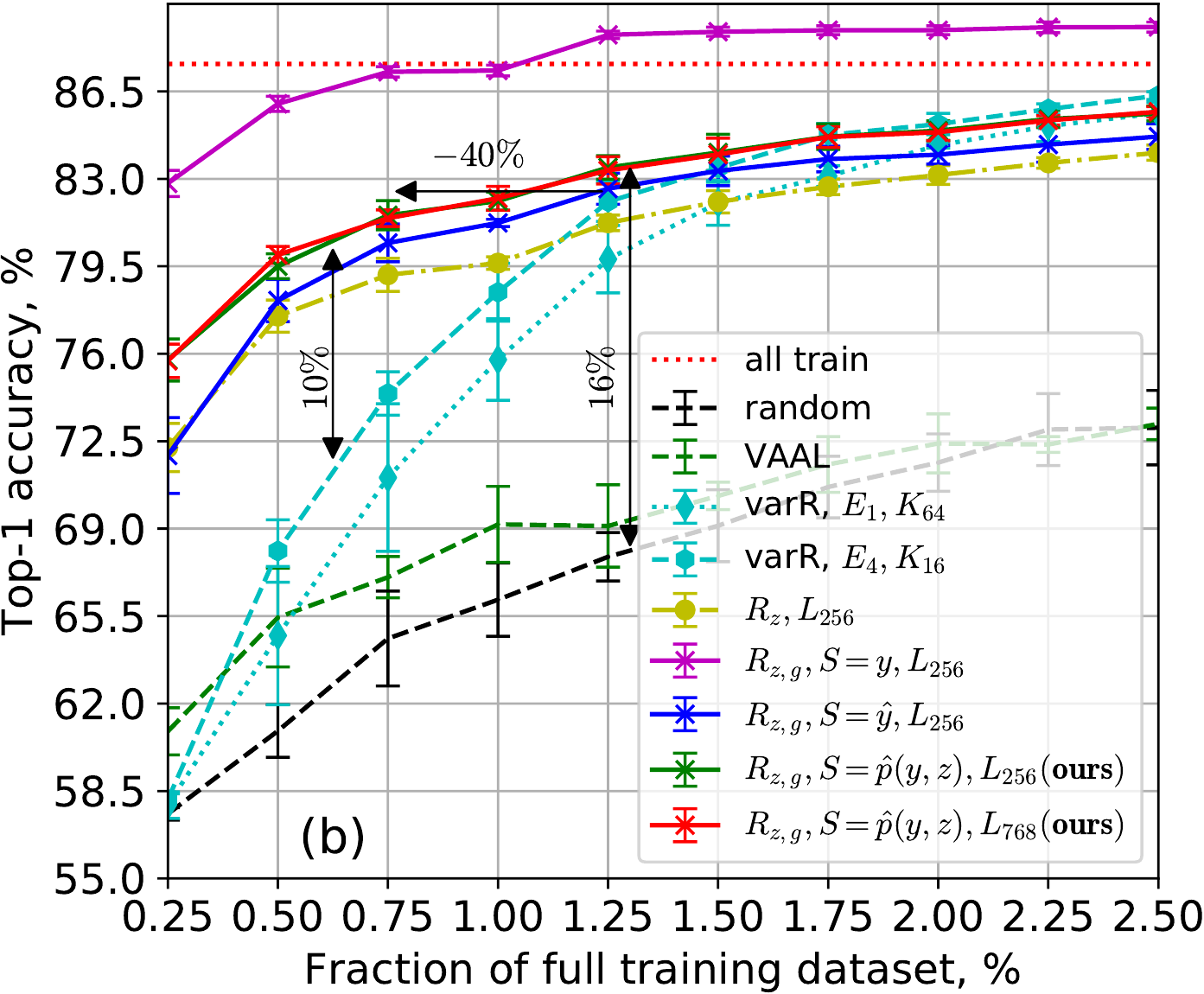} \\
		\includegraphics[width=0.32\textwidth]{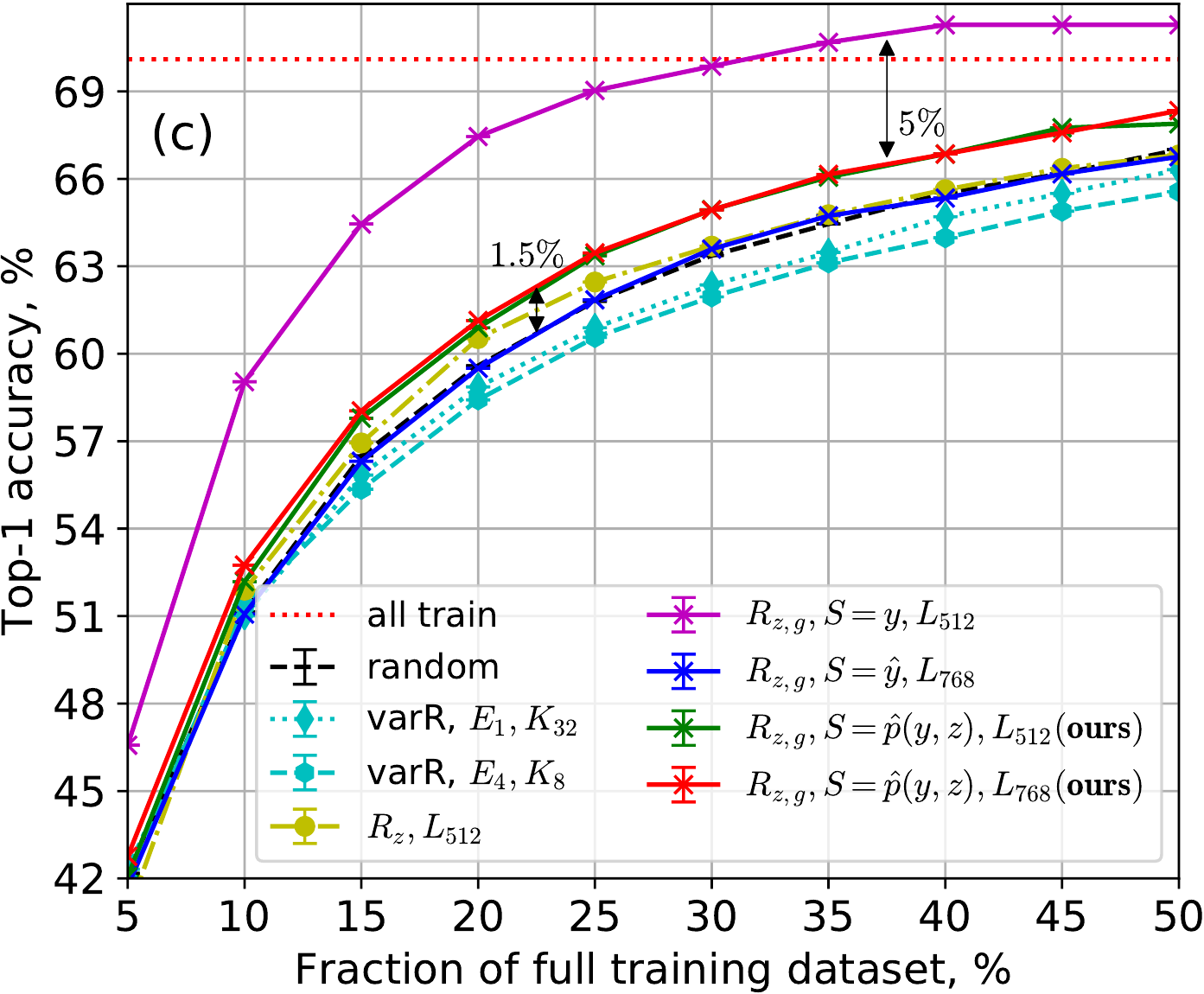} &
		\includegraphics[width=0.32\textwidth]{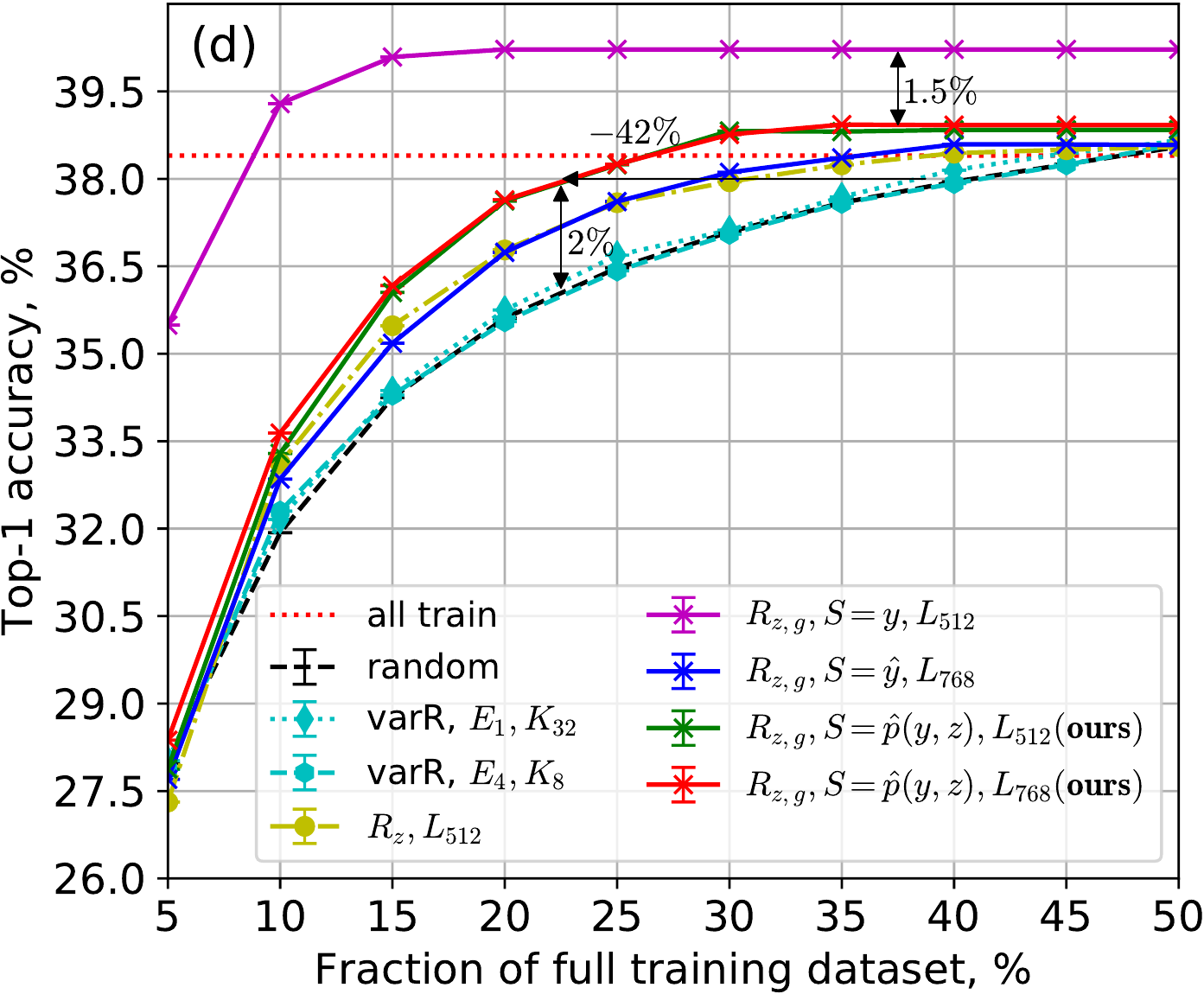}
	\end{tabular}
	\caption{SVHN test (top) and ImageNet val (bottom) accuracy: (a,c) no class imbalance and (b,d) with $100\times$ class imbalance.}
	\label{fig:svhn-fig}
\end{figure*}

\subsection{MNIST} \label{subsec:mnist}
The dataset split $|\train|$, $|\valid|$ and $|\test|$ has 50, 10 and 10 thousand images, respectively. The following hyperparameters are used: SGD, epochs=50, batch-size=25, lr=0.05, lr-decay=0.1 every 15 epochs. Descriptor length $L$ is 20 for single-scale (after \textit{conv2} output) and 80 for three-scale descriptor (\textit{conv1,2} and \textit{fc1} outputs). The selected pool size $P$ is 125 images or 0.25\% of $|\train|$.

Figure~\ref{fig:mnist-fig}(a) shows the case when the unlabeled train dataset approximates test distribution. In this setting, the uncertainty method varR performs relatively well with only 2.5\% decrease in accuracy compared to our best method $(\mR_{\vz,\vg}$, $S=\hat{p}(\vy,\vz)$, $L_{80})$ at first iterations and almost on par when $b>5$. Random sampling accuracy is only 3\% lower due to nearly uniform train distribution. VAAL~\cite{vaal} results are similar to random sampling.

A practical case with $100\times$ class imbalance is illustrated in Figure~\ref{fig:mnist-fig}(b). Our FK-based methods from~(\ref{eq:p4}) outperform PCC feature-only method from~(\ref{eq:p2}) with the increase of descriptors size $L$ and use of a better label estimation metric: $S=\hat{p}(\vy,\vz)$ vs. common $S=\hat{\vy}$. The gap between the best FK and the best uncertainty method with ensembles reaches 14\% or, equivalently, $40\%$ less labels is needed for the same accuracy. Furthermore, our method requires $EK/2=64\times$ less processing according to Table~\ref{tab:1}.

As part of ablation study, we plot in Figure~\ref{fig:mnist-fig}(a,b) a FK setup with all-true labels $(S=\vy)$. It shows the \textit{theoretical limit} of FK: no accuracy is gained without class-imbalance, while significant (3-10\%) improvement is achieved with the data bias compared to pseudo-labeling using $S=\hat{p}(\vy,\vz)$. In fact, such setup \textit{exceeds} performance of the full train dataset accuracy at the second AL iteration. Task model pretrained by rotation method is able to separate digits without supervision with exception of the last randomly initialized fully-connected layer. Hence, a single AL iteration is needed to achieve baseline result.

A set of ablation studies is presented in Figure~\ref{fig:mnist-fig}(c). First, unsupervised pretraining using rotations~\cite{gidaris} adds 7\% in accuracy when $L=20$ and 3.5\% when $L=80$ compared to random-weight initialization $(\vtheta^0_{rnd})$. Second, we compare pseudo-label estimation metrics proposed in Section~\ref{subsec:prop-pseudo}. The common $S=\hat{\vy}$ metric performs only 1\% inferior compared to MC metrics $(S=\tr{(\mC_{\vz,\vg})}~\textrm{and}~S=I(\vz;\vg))$ when $b>4$, while it requires $KD/2\times$ less processing. In our setup, MC metrics employ uniform $\pm5^{\circ}$ image rotations and Gaussian additive noise for sampling. They may require larger $K$, other sampling or go beyond the Gaussian assumption to achieve better results. For example, Kay~\etal~\cite{kay} show a tractable solution for elliptically symmetric probability model and Bachman~\etal~\cite{amdim} propose to measure mutual information across multiple scales of features. Our best metric with $S=\hat{p}(\vy,\vz)$ outperforms others by 6-7\%. Therefore, we conclude that $\mR_{\vz,\vg}$ with $S=\hat{p}(\vy,\vz)$ is a preferable approach.

\begin{figure*}[t]
	\centering
	\includegraphics[width=0.7\textwidth]{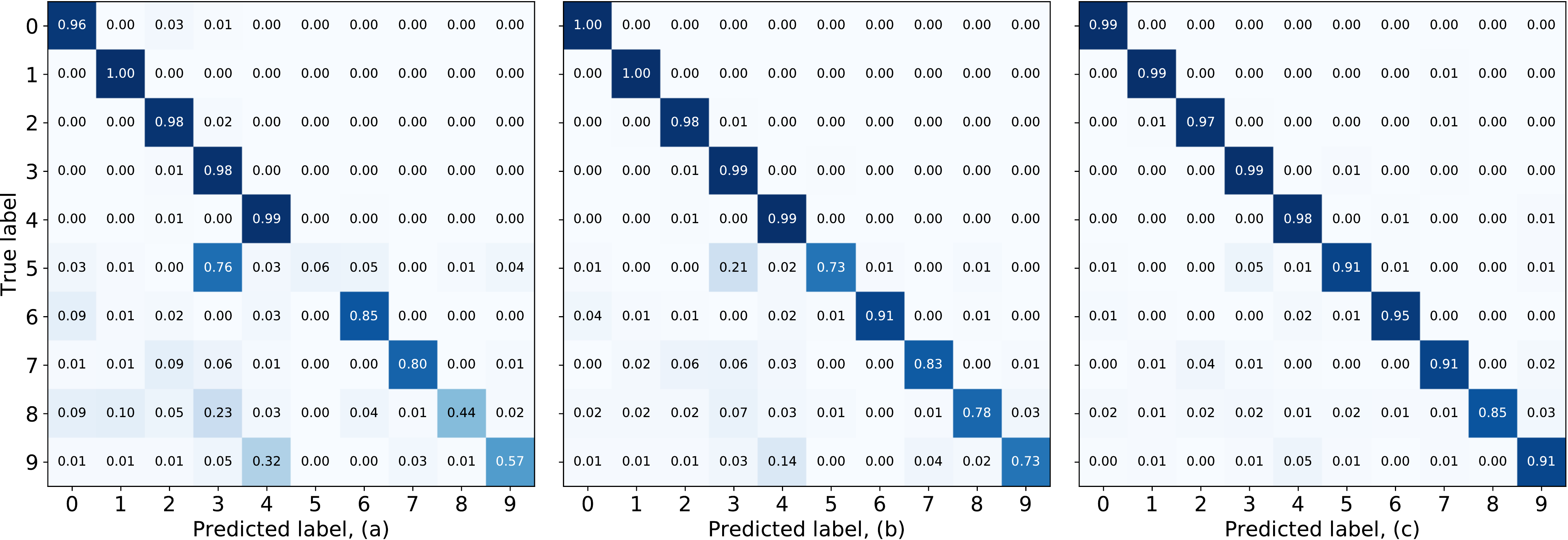}
	\includegraphics[width=0.7\textwidth]{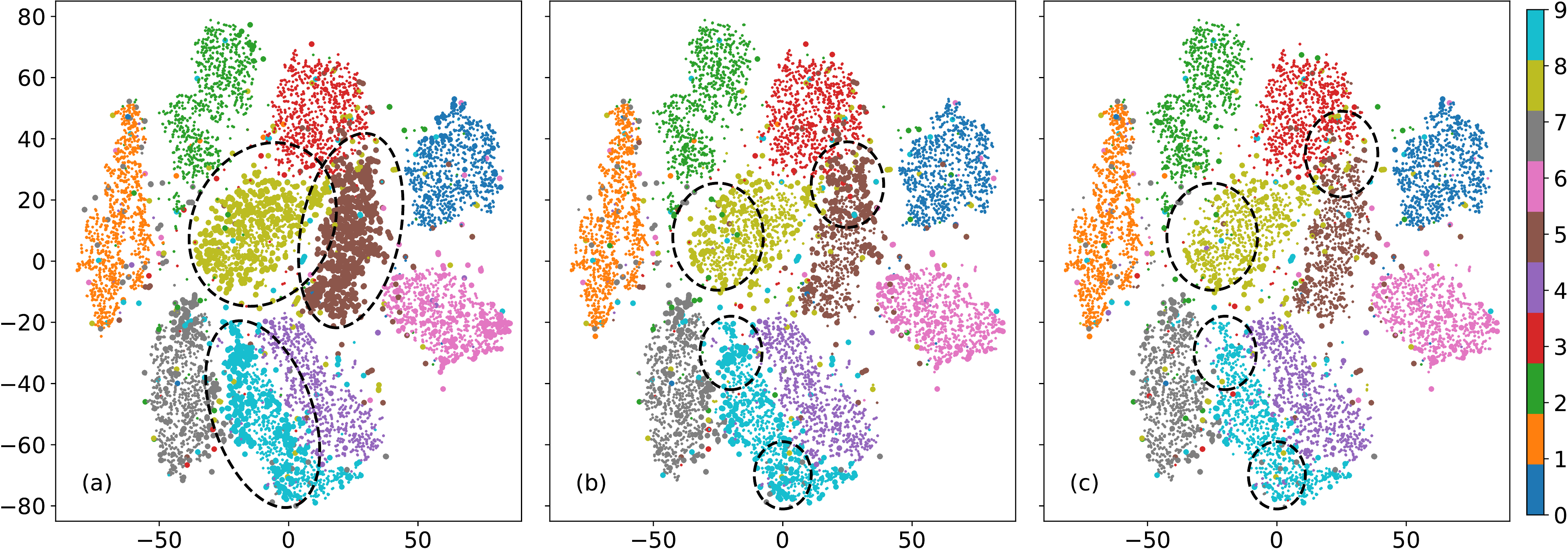}
	\caption{Confusion matrix (top) and t-SNE (bottom) of MNIST test data at AL iteration $b=3$ with 100$\times$ class imbalance for: (a) varR with $E_1$, $K_{128}$, (b) $\mR_{\vz,\vg}$, $S=\hat{p}(\vy,\vz)$, $L_{80}$ (\textbf{ours}), and (c) $\mR_{\vz,\vg}$, $S=y$, $L_{80}$. Dots and balls represent correspondingly correctly and incorrectly classified images for t-SNE visualizations. The underrepresented classes $\{5,8,9\}$ have on average 36\% accuracy for prior work~(a), while our method~(b) increases their accuracy to 75\%. The ablation configuration~(c) shows 89\% theoretical limit of our method.}
	\label{fig:mnist-cmat}
\end{figure*}

\subsection{SVHN} \label{subsec:svhn}
The dataset split $|\train|$, $|\valid|$ and $|\test|$ contains 500, 104 and 26 thousand images, respectively. Training dataset is obtained from concatenation of the original \textit{train} and \textit{extra train} datasets with total of 604,388 images. The following hyperparameters are used: SGD, epochs=35, batch-size=128, lr=0.1, lr-decay=0.1 every 15 epochs. Descriptor length $L$ is 256 for single-scale (\textit{resblock3} output) and 768 for two-scale descriptor (\textit{resblock3,4} outputs). The selected pool size $P$ is 1,250 images or 0.25\% of $|\train|$.

The gap between random sampling and our method is 3.5\% for the original and 16\% for the biased SVHN with the same amount of training data in Figures~\ref{fig:svhn-fig}(a,b). Uncertainty varR method lacks 1.5\% and 10\% in accuracy compared to ours during first AL iterations and perform on par when $b>4$. Hence, approximately 40\% of labeling can be avoided for the biased train data. Moreover, computational complexity of uncertainty methods is $32\times$ higher.

The method with PCC $(\mR_\vz)$ in Figure~\ref{fig:svhn-fig}(b) achieves 2\% and 4\% less accuracy compared to PFK $(\mR_{\vz,\vg})$ with the simplest pseudo-label estimation metric $(S=\hat{\vy})$ and our best metric $S=\hat{p}(\vy,\vz)$, respectively.

The larger descriptor size $L$ does not significantly improve accuracy in this setup. This points to importance of multi-scale extraction when, for example, spatially-localized features can be more relevant than global ones or vice versa. A parametric aggregation of feature hierarchy can lead to better results~\cite{netvlad,remap}. The latter is not trivial without labeled data, unlike our non-parametric approach.

\subsection{ImageNet} \label{subsec:imagenet}
The original dataset split $|\train|$ and $|\valid|$ has 1,200 and 50 thousand images, respectively. The following hyperparameters are used: SGD, epochs=60, batch-size=128, lr=0.1, lr-decay=0.1 at [30, 50, 57] epoch. The descriptor configuration is the same as for SVHN. The selected pool size $P$ is 64,000 images or 5\% of $|\train|$.

Figures~\ref{fig:svhn-fig}(c,d) show results for large-scale ImageNet. Uncertainty varR method underperforms without class imbalance and only a fraction of percent better than random sampling with 100$\times$ class imbalance. This could be related to lower number of samples $K$ compared to setup in~\cite{bel}, dropout setting heuristics or large number of classes. Unfortunately, it is almost infeasible to increase $K$ due to high complexity of varR, which is $16\times$ more than for our method during AL phase and $E\times$ more during retraining. For instance, the ImageNet experiment took 2.5 days for our method and 12 days for varR on a single V100 GPU.

Our best method $(\mR_{\vz,\vg}$, $S=\hat{p}(\vy,\vz)$, $L_{768})$ increases accuracy compared to prior works by 1.5\% without class imbalance and by 2\% with 100$\times$ class imbalance. The configurations with the simplest pseudo-label estimation metric $(S=\hat{\vy})$ or the ones without FK supervision gain only 1\% in accuracy. The gap between theoretically possible ImageNet result with true labels $(S=\vy)$ and our method with the estimated pseudo-labels is increasing compared to relatively small-scale 10-class MNIST in Figures~\ref{fig:mnist-fig}(a,b) and SVHN in Figures~\ref{fig:svhn-fig}(a,b). It indicates that a more accurate pseudo-label metric may improve results even more. While our absolute accuracy improvement is 2\%, it leads to 42\% less annotations with the same accuracy.

\subsection{Qualitative visualizations}
\label{subsec:tsne}
To demonstrate improvement of AL behavior, we calculate confusion matrices and t-SNE~\cite{tsne} clusters. We use the same experimental setup as in Figure~\ref{fig:mnist-fig}(b) with class imbalance ratio of 100 and analyze MNIST test dataset after the third AL iteration $(b=3)$. Figure~\ref{fig:mnist-cmat} presents results for the following configurations: (a) varR $(E_1$, $K_{128})$ and the proposed $(\mR_{\vz,\vg},L_{80})$ with (b) pseudo-labels $(S=\hat{p}(\vy,\vz))$ and (c) true-labels $(S=\vy)$ for ablation study. 

The class-imbalanced digits $\{5\ldots9\}$ are heavily misclassified in Figure~\ref{fig:mnist-cmat}(a). It visually confirms quantitative result from Section~\ref{subsec:mnist} that uncertainty methods fail to identify relevant training data clusters. Those methods can only capture so called \textit{epistemic} uncertainty which is uncertainty over DNN parameters instead of uncertainty about data.

Figures~\ref{fig:mnist-cmat}(b,c) show results of the FK-supervised methods with the estimated pseudo-labels and true-labels. Compared to Figure~\ref{fig:mnist-cmat}(a) the class-imbalanced digits are significantly better classified, specifically, the centers of clusters "5", "8" and "9", whose average accuracy increased from only 36\% to 75\%. This result indicates the ability of self-supervised FK to find long-tails of distribution using our acquisition function~(\ref{eq:p4}).

The far edges of the imbalanced clusters that intersect with other digit clusters still experience some irregular densities of misclassified examples in Figure~\ref{fig:mnist-cmat}(b) due to imperfect pseudo-labeling. The t-SNE setup with all-true labels in Figure~\ref{fig:mnist-cmat}(c) improves on those edges and achieves 89\% accuracy. Clearly, it is the most difficult to separate very similar intersecting examples from different classes. As a potential future direction, this problem might be addressed by a better feature separation or using adversarial training.

\section{Conclusions}
\label{sec:conclusion}
We formulated the optimal acquisition function for AL with realistic assumptions about data biases and continuous updates after field trials. We introduced low-complexity non-parametric AL method that minimizes distribution shift between train and validation datasets using self-supervised FK and several novel pseudo-label estimators. According to ablation studies, unsupervised pretraining further improved our approach. The conducted image classification experiments showed that our method results in at least $40\%$ less labeling for biased data compared to prior works while requiring a factor of $10$ less processing.

{\small
\bibliographystyle{ieee_fullname}
\bibliography{paper_2499}
}

\appendix
\section{Problem Statement for Biased Datasets}
\label{sec:atheory}

Using definitions of $q(\vx,\vy)$ and $p(\vx,\vy | \vtheta)$,~(\ref{eq:t2}) can be analytically derived as
\begin{equation*} \label{eq:at2}
\begin{split}
D_{KL}&(Q_{\vx,\vy} \| P_{\vx,\vy}(\vtheta)) = \\
&\int \int q(\vy | \vx) q(\vx) \log \frac{q(\vy | \vx) q(\vx)}{p(\vy | \vx, \vtheta) q(\vx)} d\vy d\vx = \\
&\int q(\vx) \int q(\vy | \vx) \log \frac{q(\vy | \vx)}{p(\vy | \vx, \vtheta)} d\vy d\vx = \\
&\sE_{Q_\vx}[ D_{KL}(Q_{\vy|\vx} \| P_{\vy|\vx}(\vtheta) ) ].
\end{split}
\end{equation*}

Assuming that $Q_{\vy|\vx}$ can be replaced by empirical $\hat{Q}_{\vy|\vx}$ and $\vy=\vone_d\in\mathbb{R}^{D}$ is one-hot vector with only $d$th class not equal to zero,~(\ref{eq:t4}) can be derived as
\begin{equation*} \label{eq:at4}
\begin{split}
\mathcal{L}(\vtheta) & = \frac{1}{N^b} \sum_{i \in \sN^b} [ D_{KL}(Q_{\vy_i|\vx_i} \| P_{\vy_i|\vx_i}(\vtheta)) ] = \\
& = \frac{1}{N^b} \sum_{i \in \sN^b} \sum_{d=1}^{D} \vone_d(i) \log \frac{\vone_d(i)}{p(\vy_i|\vx_i, \vtheta)} = \\
& -\frac{1}{N^b} \sum_{i \in \sN^b} \log p(\vy_i|\vx_i, \vtheta).
\end{split}
\end{equation*}

\section{Relationship between $D_{KL} (P_{\vz}^{\mathrm{v}} \| P_{\vz})$ and Fisher Information}
\label{sec:aconn}
Using the sufficiency property~\cite{achille}, we approximate our optimal acquisition function~(\ref{eq:t5}) using the distributions of learned representations $\vz$ as
\begin{equation*} \label{eq:ap22}
\mathcal{R}_{opt}(b,P) = \argmin_{\mathcal{R}(b,P)} D_{KL}(\hat{P}_{\vz}^{\mathrm{v}} \| \hat{P}_{\vz}),
\end{equation*}

Then, a \textit{connection} between the main task~(\ref{eq:t2}) and $D_{KL}(P_{\vz}^{\mathrm{v}} \| P_{\vz})$ minimization in~(\ref{eq:p22}) via Fisher information can be derived with respect to small perturbations in $\vtheta$. Assuming that the task model minimizes distribution shift in~(\ref{eq:t2}) every backward pass as
\begin{equation*} \label{eq:ap23}
p^{\mathrm{v}}(\vz|\vtheta) = p(\vz|\vtheta) + \Delta p,
\end{equation*}
where $\Delta p = \Delta \vtheta \frac{\partial p(\vz|\vtheta)}{\partial \vtheta}$ and $\Delta \rightarrow 0$.

By substituting~(\ref{eq:p23}), the expanded form of $D_{KL}(P_{\vz}^{\mathrm{v}} \| P_{\vz})$ can be written as
\begin{equation*} \label{eq:ap25}
\begin{split}
& D_{KL} (P_{\vz}^{\mathrm{v}} \| P_{\vz}) = \int \left( p(\vz|\vtheta) + \Delta p \right) \log \frac{p(\vz|\vtheta) + \Delta p}{p(\vz|\vtheta)} d\vz = \\
& \int \left( p(\vz|\vtheta) + \Delta p \right) \log \left(1 + \frac{\Delta p}{p(\vz|\vtheta)} \right) d\vz.
\end{split}
\end{equation*}

Using the Taylor series of natural logarithm, this can be approximated by
\begin{equation*} \label{eq:ap26}
\begin{split}
& D_{KL} (P_{\vz}^{\mathrm{v}} \| P_{\vz}) \approx \int \left( p(\vz|\vtheta) + \Delta p \right) \times \\
& \left( \frac{\Delta p}{p(\vz|\vtheta)} - \frac{(\Delta p)^2}{2(p(\vz|\vtheta))^2} \right) d\vz = \int \Delta p d\vz + \\
& \frac{1}{2} \int \left( \frac{\Delta p}{p(\vz|\vtheta)} \right)^2 p(\vz|\vtheta) d\vz - \int \frac{(\Delta p)^3}{2p(\vz|\vtheta)^2} d\vz,
\end{split}
\end{equation*}
where the first term using the definition of $\Delta p$ is equal to zero and the third $\mathcal{O}(\Delta \vtheta^3) \rightarrow 0$.

By substituting $\Delta p$ and rewriting vector $\vtheta$ as a discrete sum, the term
\begin{equation*} \label{eq:ap27}
\begin{split}
& \frac{\Delta p}{p(\vz|\vtheta)} \approx \sum_i \frac{\partial \log p(\vz|\vtheta)}{\partial \vtheta_i} \Delta \vtheta_i.
\end{split}
\end{equation*}

Using this approximation, the final form of~(\ref{eq:p22}) can be obtained as
\begin{equation*} \label{eq:ap28}
\begin{split}
& \mathcal{R}_{opt}(b,P) = \argmin_{\mathcal{R}(b,P)} D_{KL} (P_{\vz}^{\mathrm{v}} \| P_{\vz}) \\
& \approx \argmin_{\mathcal{R}(b,P)} \sum_{m,n} \mathcal{\mI}_{m,n} \Delta \vtheta_m \Delta \vtheta_n \approx \argmin_{\mathcal{R}(b,P)} \Delta \vtheta^T \mathcal{\mI} \Delta \vtheta,
\end{split}
\end{equation*}
where $\mathcal{\mI} = \sE_{P_\vz} \left[ \vg(\vtheta) \vg(\vtheta)^T \right]$ is a Fisher information matrix and $\vg(\vtheta) = \frac{\partial \log p(\vz|\vtheta)}{\partial \vtheta}$ is a Fisher score with respect to $\vtheta$.

\section{Practical Fisher Kernel for DNNs}
\label{sec:adnn}
Using the chain rule for a DNN layer $(\tilde{\vz}^j_i = \vtheta^T \vz^j_i = \vtheta^T \sigma(\tilde{\vz}^{j-1}_i))$ with $\sigma(\cdot)$ nonlinearity, Jacobian of interest can be simplified as follows
\begin{equation*} \label{eq:ap30}
\frac {\partial L(\vy_i, \hat{\vy}_i)} {\partial \vtheta} = \frac {\partial L(\vy_i, \hat{\vy}_i)} {\partial \tilde{\vz}_i} \frac {\partial \tilde{\vz}_i} {\partial \vtheta} = \frac {\partial L(\vy_i, \hat{\vy}_i)} {\partial \tilde{\vz}_i} \vz^T_i = \vg_i \vz^T_i,
\end{equation*}
where $\vtheta\in\mathbb{R}^{L \times L}$, $\vz_i\in\mathbb{R}^{L \times 1}$, and $\vg_i\in\mathbb{R}^{L \times 1}$.

Then, approximation of FK in~(\ref{eq:p32}) for $\vg_i(\vtheta) = \mathrm{vec} (\partial L(\vy_i, \hat{\vy}_i) / \partial \vtheta)\in\mathbb{R}^{L^2 \times 1}$ can be derived as
\begin{equation*} \label{eq:ap3}
\begin{split}
& \emR_{z,g}(\vz_m, \vz_n) = \vg_m(\vtheta)^T \mathcal{\mI}^{-1} \vg_n(\vtheta) \stackrel{\text{PFK}}{\approx} \vg_m(\vtheta)^T \vg_n(\vtheta) = \\
& \mathrm{vec} \left( \frac{\partial L(\vy_m, \hat{\vy}_m)}{\partial \tilde{\vz}_m} \vz^T_m  \right)^T \mathrm{vec} \left( \frac{\partial L(\vy_n, \hat{\vy}_n)}{\partial \tilde{\vz}_n} \vz^T_n  \right) = \\
& \mathrm{vec} \left( \vg_m  \vz^T_m  \right)^T \mathrm{vec} \left( \vg_n  \vz^T_n  \right) = [g^1_m \vz_m, g^2_m \vz_m, \ldots, g^L_m \vz_m]^T \times \\
& [g^1_n \vz_n, g^2_n \vz_n, \ldots, g^L_n \vz_n] = \vz_m^T \vz_n \sum^L_l g^l_m g^l_n = \vz_m^T \vz_n \vg_m^T \vg_n.
\end{split}
\end{equation*}

\end{document}